\def\year{2022}\relax
\documentclass[letterpaper]{article} 
\usepackage{aaai22}  
\usepackage{times}  
\usepackage{helvet}  
\usepackage{courier}  
\usepackage[hyphens]{url}  
\usepackage{graphicx} 
\usepackage{natbib}  
\usepackage{caption} 
\DeclareCaptionStyle{ruled}{labelfont=normalfont,labelsep=colon,strut=off} 
\usepackage{algorithm}
\usepackage{algorithmic}

\usepackage{amssymb}
\usepackage{amsmath}
\usepackage{booktabs}
\usepackage{multirow}
\usepackage{subcaption}

\usepackage{newfloat}
\usepackage{listings}
\floatstyle{ruled}
\newfloat{listing}{tb}{lst}{}
\floatname{listing}{Listing}
\author {
    Yitao Liu,\equalcontrib
    Tianxiang Sun,\equalcontrib
    Xipeng Qiu,\thanks{Corresponding author: Xipeng Qiu (xpqiu@fudan.edu.cn).}
    Xuanjing Huang
}
\affiliations {
     School of Computer Science, Fudan University\\
     Shanghai Key Laboratory of Intelligent Information Processing, Fudan University\\
    \{yitaoliu20, txsun19, xpqiu, xjhuang\}@fudan.edu.cn
}

\title{Learning to Teach with Student Feedback}

\usepackage{bibentry}

\begin{document}

\maketitle

\begin{abstract}
Knowledge distillation (KD) has gained much attention due to its effectiveness in compressing large-scale pre-trained models. In typical KD methods, the small student model is trained to match the soft targets generated by the big teacher model. However, the interaction between student and teacher is one-way. The teacher is usually fixed once trained, resulting in static soft targets to be distilled. This one-way interaction leads to the teacher's inability to perceive the characteristics of the student and its training progress. To address this issue, we propose Interactive Knowledge Distillation (IKD), which also allows the teacher to learn to teach from the feedback of the student. In particular, IKD trains the teacher model to generate specific soft target at each training step for a certain student. Joint optimization for both teacher and student is achieved by two iterative steps: a course step to optimize student with the soft target of teacher, and an exam step to optimize teacher with the feedback of student. IKD is a general framework that is orthogonal to most existing knowledge distillation methods. Experimental results show that IKD outperforms traditional KD methods on various NLP tasks.
\end{abstract}

\section{Introduction}

Large-scale pre-trained language models (PLMs) such as GPT~\cite{radford2019language}, BERT~\cite{devlin-etal-2019-bert}, RoBERTa~\cite{liu2019roberta} have achieved significant improvement on various NLP tasks. Despite their power, they are computationally expensive due to their enormous size, which limits their deployment in real-time scenarios.

As an effective technique to tackle this problem, Knowledge Distillation (KD)~\cite{Bucila2006Compression,Hinton2015Distilling} has gained much attention in the community. In common KD methods for compressing large-scale PLMs, the original large-scale PLM often serves as the teacher model, which is first trained on the downstream task then uses its generated soft targets to teach the student model, which is usually a smaller PLM. Previous works such as DistilBERT~\cite{sanh2019distilbert}, BERT-PKD~\cite{sun-etal-2019-patient}, MobileBERT~\cite{Sun2020Mobile}, and TinyBERT~\cite{jiao-etal-2020-tinybert} have explored informative features that can be distilled, including the output logits, word embeddings, hidden states, attention maps, etc.

\begin{figure}[t]
    \centering
    \includegraphics[width=.7\linewidth]{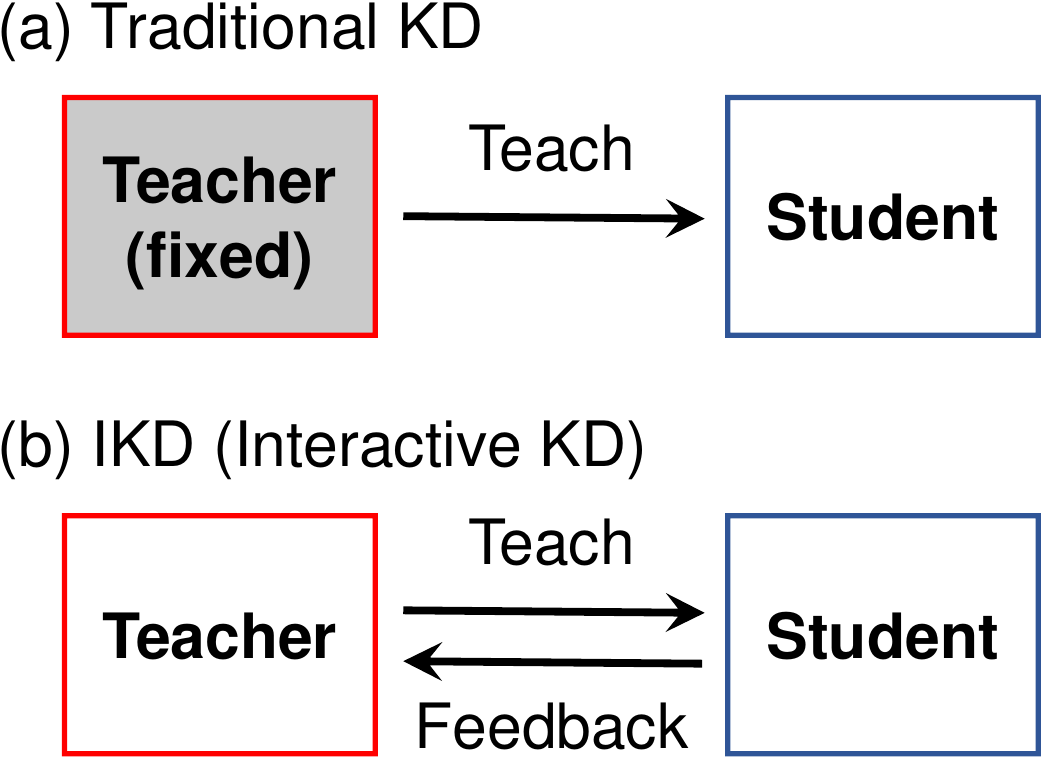}
    \caption{Comparison between traditional KD methods and our proposed IKD. (a) Traditional KD methods use an one-way interaction between teacher and student while the teacher model is static during teaching student. (b) IKD builds a co-interactive channel to achieve joint training of both teacher and student.}
    \label{fig:comp}
\end{figure}

However, in these existing methods the interaction between teacher and student remains one-way. The teacher model is usually fixed once trained on the downstream data, resulting in static soft targets regardless of the characteristics of the student and its training progress. Though there are works that insert a scheduled temperature into the Softmax function to control the smoothness of the soft targets such as Annealing-KD~\cite{Jafari2021Annealing}, the design of the temperature and its schedule does not utilize the feedback from the student and highly depends on expert experience. 

In this paper, we propose the Interactive Knowledge Distillation (IKD) to implement a bidirectional interaction between teacher and student. IKD allows the teacher to learn to teach based on the feedback from the student. In particular, the teacher model is trained to generate specific soft targets at each training step with the help of meta-learning, especially MAML~\cite{DBLP:conf/icml/FinnAL17}. Figure~\ref{fig:comp} illustrates main difference between our proposed IKD and traditional KD methods.

The central idea of IKD is to make the student model generalize well (like achieving a lower loss) on a batch of unseen samples after learned from the teacher model, while the measurable performance on unseen samples is back-propagated to the teacher model in order to update its teaching strategy. More specifically, IKD consists of two update steps: \textit{course} step and \textit{exam} step. \textbf{In course step}, the student is trained to match the soft targets generated by the teacher on a batch of data called course data. \textbf{In exam step}, the student after one gradient step is evaluated on another batch of data called exam data, the cross entropy of the student on the exam data provides meta-gradients to be back-propagated to the teacher model such that the teacher model could get updated to generate better soft targets. By iteratively conducting the two steps, we can continually improve the student model via the soft targets generated by the teacher, and improve the teacher model via the feedback generated by the student.

We conduct experiments on various NLP tasks. Experimental results on GLUE benchmark demonstrate that IKD consistently outperforms vanilla knowledge distillation. Further analysis is then conducted to shed some light on the dynamic soft targets and the student feedback. 

To sum up, our contributions are as follows:
\begin{itemize}
    \item We propose a co-interactive method for teacher-student framework, namely Interactive Knowledge Distillation (IKD), which allows the teacher model to learn to generate specific soft targets at each training step.
    \item We take an approximation to convert the iterative optimization into a joint optimization, which is more efficient for training.
    \item Our proposed IKD is orthogonal to most existing knowledge distillation works that distill different feature sets, in which we empirically evaluated the effectiveness of IKD based on vanilla knowledge distillation.
\end{itemize}

\section{Method}
\label{sec:method}
\begin{figure*}[thbp]
    \centering
    \includegraphics[width=\linewidth]{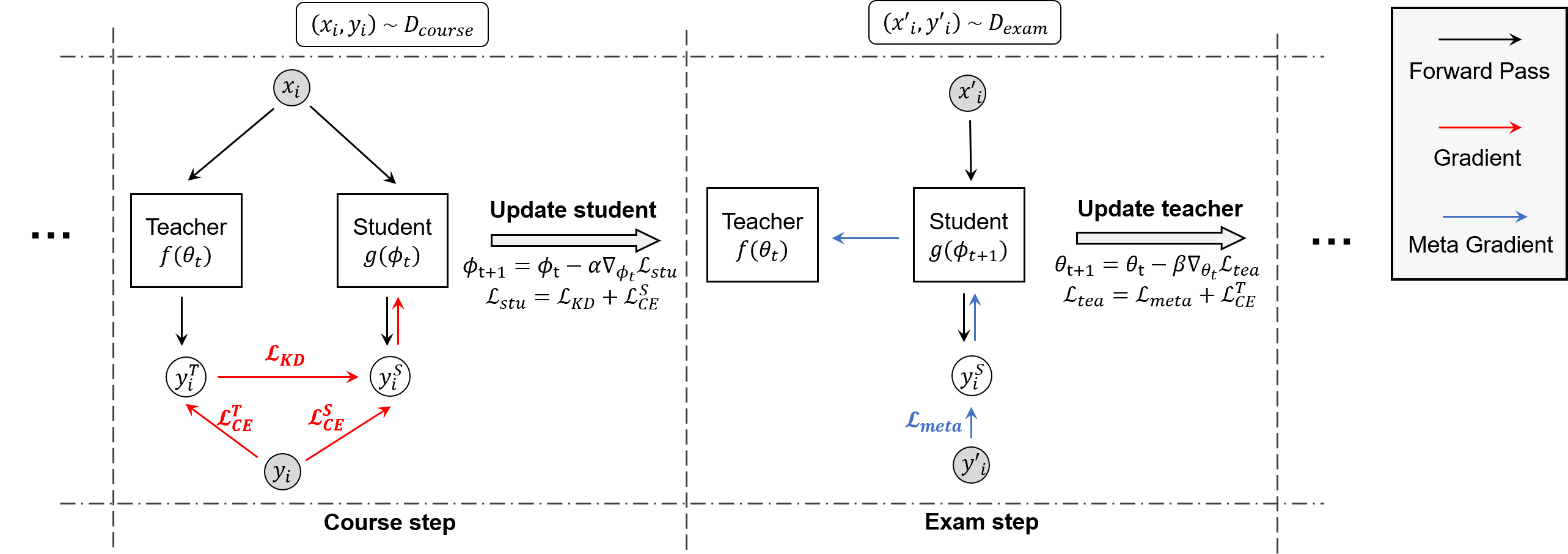}
    \caption{Illustration of our proposed Interactive Knowledge Distillation (IKD). IKD consists of two iterative steps: (a) Course step to update student with the soft targets generated by the teacher, (b) Exam step to update teacher with the feedback (i.e. cross entropy on the exam data) produced by the student. By iterating the course step and the exam step, both of student and teacher can be optimized.}
    \label{fig:illus}
\end{figure*}

The intuition behind our framework is straight-forward: First, the teacher teaches the \textit{course} to the student who then updates its knowledge according to the course. Second, the student takes \textit{exams} and produces scores for the teacher to adjust its teaching strategy. Such two steps are common in real-world education.

Based on the intuition above, IKD can be formulated in the context of machine learning as follows. Denote the teacher model and the student model as \(f\) and \(g\) respectively. Assume teacher model \(f\) is parameterized by \(\theta\), student model \(g\) is parameterized by \(\phi\). In the course step, we optimize the student model \(f(\theta)\) by vanilla knowledge distillation. The student model is trained to match the soft targets generated by the teacher model on a batch of data drawn from the course data set \(D_{course}\). In the exam step, we evaluate the student model on a batch of data drawn from the exam data set \(D_{exam}\). The test score of the exam, which can be instantiated as cross entropy, provides meta gradients to optimize the teacher model \(g(\phi)\). The overall illustration of our method is depicted by Figure~\ref{fig:illus}.

\subsection{Course Step: Student Optimization by Vanilla Knowledge Distillation}
\label{sec:Vanilla KD}
In our setting, the teacher model \(f(\theta)\) is implemented as a deep encoder such as BERT~\cite{devlin-etal-2019-bert}, and the student model \(g(\phi)\) is implemented as a lightweight encoder such as a Transformer~\cite{vaswani2018attention} encoder with fewer layers.

Given a training sample \((x_i,y_i)\) drawn from the course data set \(D_{course}=\{x_i,y_i\}_{i=1}^N\), the teacher model computes a contextualized embedding \(h_i^T=f(x_i;\theta)\)\footnote{We use the \texttt{[CLS]} token embedding of the last layer as the sentence representation.}, which is then fed into a Softmax layer to obtain the probability for each category, i.e. the soft targets, that is \(y_i^T=\mathrm{Softmax}(W^Th_i^T)\), where \(W^T\) is a learnable parameter matrix and \(W^Th_i^T\) is the logits. For simplicity, \(W^Th_i^T\) is abbreviated to \(z_i^T\). In practice, the temperature is often introduced as a hyper-parameter to control the smoothness of the soft targets,
\begin{equation}
y_i^T = \mathrm{Softmax}(z_i^T/T) = \frac{e^{z_{i,c}^T/T}}{\sum_{c^\prime=1}^C{e^{z_{i,c^\prime}/T}}},
\end{equation}
where \(C\) denotes the number of classes, \(c\) is the correct class.

Similarly, we can obtain the output of the student model \(y_i^S=\mathrm{Softmax}(W^Sh_i^S)\), where \(h_i^S=g(x_i;\phi)\). In vanilla knowledge distillation, the KL divergence of the teacher's prediction and the student's prediction should be minimized,
\begin{equation}
\mathcal{L}_{KD} = \mathrm{KL}(y^T, y^S) = -\frac{1}{N}\sum_{i=1}^{N}\sum_{j=1}^{C}y_{i,j}^T\log \frac{y_{i,j}^S}{y_{i,j}^T},
\end{equation}
where \(\mathrm{KL}\) means Kullback-Leibler divergence.

In practice, the cross entropy between the student's prediction and the ground truth is also incorporated to be minimized, i.e.,
\begin{equation}
\label{eq:L_stu}
\mathcal{L}_{CE}^S = -\frac{1}{N}\sum_{i=1}^{N}\sum_{j=1}^{C}y_{i,j}\log y_{i,j}^S.
\end{equation}

Thus the overall loss function for the student model can be written as:
\begin{equation}
\label{eq:L_tildeA}
\mathcal{L}_{stu} = \lambda \mathcal{L}_{KD} + (1-\lambda)\mathcal{L}_{CE}^S,
\end{equation}
where \(\lambda\) is a hyper-parameter to balance the knowledge distillation loss and the cross entropy loss.

The student model is then updated by taking one gradient descent step with the loss function above, i.e.
\begin{equation}
\phi _{t+1} = \phi _t - \alpha \nabla _{\phi_t} \mathcal{L}_{stu},
\end{equation}
where \(\alpha\) is the learning rate of the student model.

\subsection{Exam Step: Teacher Optimization by Student's Feedback}
It is expected that the student with the post-update parameters will generalize well on unseen samples. Therefore, the updated student is then evaluated on another batch of samples drawn from the exam data set \(D_{exam}=\{x_i^\prime, y_i^\prime\}_{i=1}^{M}\). The performance of the student is measured by the cross entropy loss on the batch of exam data, which is called \textit{meta loss} since it is calculated on the post-update parameter \(\phi _{t+1}\) thus provides meta gradient w.r.t. the teacher's parameters.
\begin{align}
\mathcal{L}_{meta}
&= \mathrm{CE}(y^\prime, y_{t+1}^{\prime S}) \\ 
&= -\frac{1}{M}\sum_{i=1}^{M}\sum_{j=1}^{C}y_{i,j}^\prime\log y_{i,j}^{\prime S}(\phi _{t+1}).
\end{align}



\begin{figure}[t]
    \centering
    \includegraphics[width=.9\linewidth]{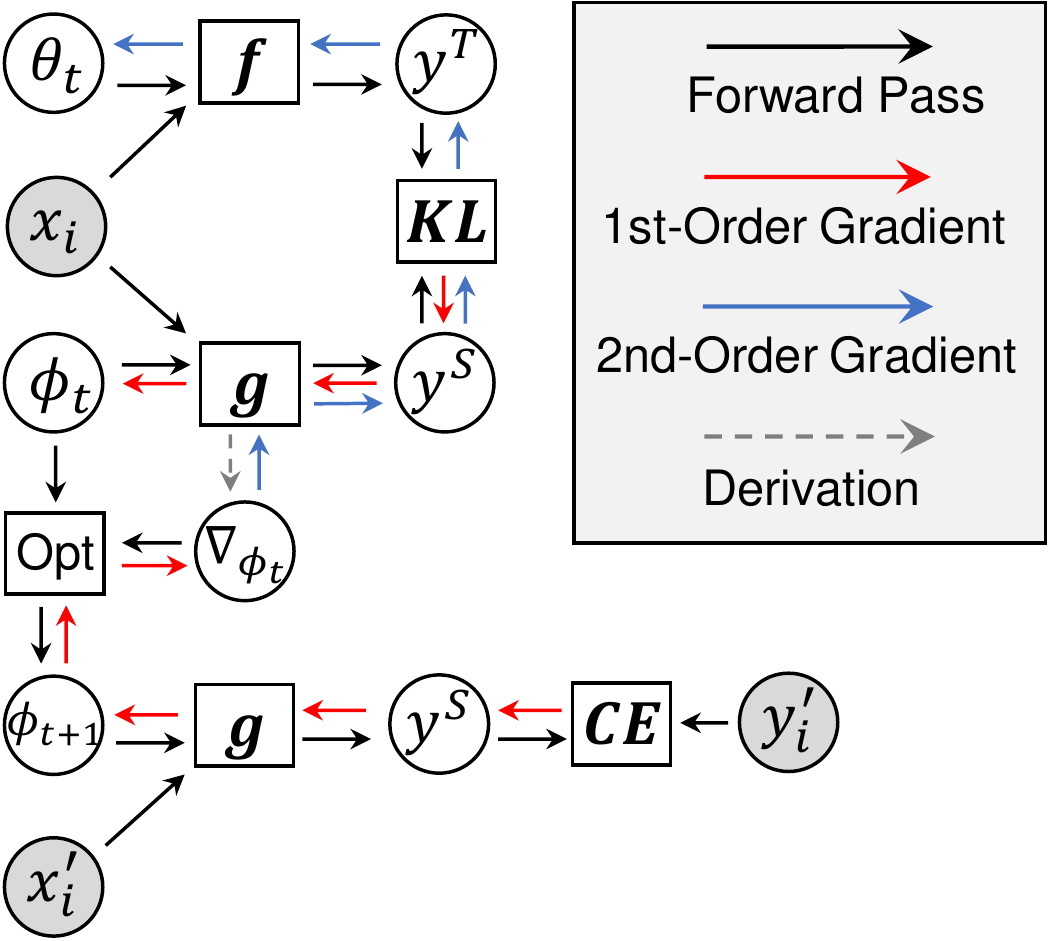}
    \caption{Computational graph to show how gradients are back-propagated into the teacher's parameters. Circles represent variable and squares represent operation. "Opt" stands for the optimizer, which is one gradient descent step. For simlicity, we omit the cross entropy loss for the student \(\mathcal{L}_{CE}^S\) and for the teacher \(\mathcal{L}_{CE}^T\) in the course step.}
    \label{fig:gradient}
\end{figure}

Our goal is to minimize \(\mathcal{L}_{meta}\) by optimizing the teacher model, therefore the gradient of \(\mathcal{L}_{meta}\) is w.r.t. the teacher's parameters \(\theta\), so is a second-order gradient (gradient of gradient). To explicitly show this, we can re-write \(y_{i,j}^{\prime S}(\phi _{t+1})\) in the above equation as:
\begin{equation}
    y_{i,j}^{\prime S}(\phi _{t+1}) = y_{i,j}^{\prime S}(\phi _{t} - \alpha \nabla_{\phi_t}\mathcal{L}_{stu}(\theta_t)),
\end{equation}
in which we know that the gradient can flow into \(\theta_t\) through the knowledge distillation in the course step. Figure~\ref{fig:gradient} shows the computational graph of the forward pass and the backward pass.

Although it is computationally available by conducting an additional backward pass under current deep learning libraries that support automatic differentiation, we use the first-order approximation for its efficiency~\cite{DBLP:conf/icml/FinnAL17,nichol2018fomaml}.

In the context of meta-learning, the teacher can be regarded as the meta-learner, while \(D_{course}\) and \(D_{exam}\) can be analogous to support set and query set respectively.

Also, the standard cross entropy loss between the teacher's prediction and the ground truth can be incorporated,
\begin{equation}
\label{eq:L_tea}
\mathcal{L}_{CE}^T = -\frac{1}{N}\sum_{i=1}^{N}\sum_{j=1}^{C}y_{i,j}\log y_{i,j}^T.
\end{equation}

Thus, the overall loss function for the teacher model is
\begin{equation}
\label{eq:L_tildeB}
\mathcal{L}_{tea} = \gamma \mathcal{L}_{meta} + (1-\gamma)\mathcal{L}_{CE}^T,
\end{equation}
where \(\gamma\) is another hyper-parameter to balance the meta loss and the cross entropy loss.

Thus, the teacher model can also be optimized by taking gradient descent:
\begin{equation}
\theta _{t+1} = \theta _t - \beta \nabla _{\theta_t} \mathcal{L}_{tea},
\end{equation}
where \(\beta\) is the learning rate of the teacher model.

Putting the student optimization in course step and the teacher optimization in exam step together, the overall loss function of IKD is defined as
\begin{equation}
\mathcal{L}_{IKD} = \mathcal{L}_{stu} + \mathcal{L}_{tea}.
\end{equation}
By iterating the course step and the exam step, the student model learns from the soft targets generated by the teacher while the teacher model learns from the feedback generated by the student.

\subsection{Discussion and Implementation}
\label{sec:Implementation Details}
To take a closer look, in this section we further analyze the expanded form of the meta-gradient, which provides insights into the \textit{student's feedback} and sheds light on how to take an approximation to achieve efficient training.

According to the chain rule, we can unfold the meta gradient \(\nabla _{\theta_t} \mathcal{L}_{meta}\) as follows\footnote{For the simplicity of derivation, we set \(N\) and \(M\) to 1, and omit the constant scalar \(\lambda\).},
\begin{equation}
\label{eq:L_B}
\begin{aligned}
\nabla _{\theta_t} \mathcal{L}_{meta} 
&= \nabla _{\phi_{t+1}} \mathcal{L}_{meta} \cdot \nabla _\theta {\phi_{t+1}}  \\
&= \nabla _{\phi_{t+1}} \mathcal{L}_{meta} \cdot (-\alpha \nabla _\theta {\nabla _{\phi_t} {\mathcal{L}_{stu}}})  \\ 
&= \alpha \nabla _{\phi_{t+1}} \mathcal{L}_{meta} \cdot \nabla _\theta {\nabla _{\phi_t} \log y_{t}^S \cdot y_{t}^T}  \\ 
&= \underbrace{\alpha \nabla _{\phi_{t+1}} \mathcal{L}_{meta} \cdot \nabla _{\phi_{t}} \log y_{t}^S}_{\text{feedback}} \cdot \nabla _{\theta_{t}} y_{t}^T.
\end{aligned}
\end{equation}

Note that \(\nabla _{\phi_{t+1}} \mathcal{L}_{meta}\) is the Jacobi matrix of shape \(|\phi|\) and \(\nabla _{\phi_{t}} \log y_{t}^S\) is of shape \(|\phi| \times C\).\footnote{For matrix multiplication, \(\nabla _{\phi_{t}} \log y_{t}^S\) should be in the transpose form, which is omitted here for the simplicity.} Therefore, \(\nabla _{\phi_{t+1}} \mathcal{L}_{meta} \cdot \nabla _{\phi_{t}} \log y_{t}^S\) is actually a weighting vector of shape \(C\) to adjust the gradient of teacher's prediction w.r.t. its parameters \(\nabla _{\theta_{t}} y_{t}^T\). It is easy to find that the weighting vector only depends on the student's gradient, therefore it is exactly the student's feedback.

The behavior relationship between the weighting vector and the teacher's prediction is analyzed. It is worth noticing that the weighting vector should be detached from the computational graph to forbid the gradient flow into the student. By this, the meta loss will only contribute to the update of the teacher's parameters.

In practice, the difference between \(\phi_{t}\) and \(\phi_{t+1}\) is usually very small, which motivates us to take an approximation \({\phi_{t+1}} \approx {\phi_{t}}\) such that \({\nabla _{\phi_{t+1}} \mathcal{L}_{meta}} \approx {\nabla _{\phi_{t}} \mathcal{L}_{meta}}\). By this approximation, Eq. (\ref{eq:L_B}) can be simplified as
\begin{equation}
\label{eq:L_B approx}
\nabla _{\theta_t} \mathcal{L}_{meta} \approx{ \alpha \nabla _{\phi_{t}} \mathcal{L}_{meta} \cdot \nabla _{\phi_{t}} \log y_{t}^S \cdot \nabla _{\theta_{t}} y_{t}^T}.
\end{equation}

Besides, note that \(\nabla _{\phi_{t}} \mathcal{L}_{meta}\), \(\nabla _{\phi_{t}} \log y_{t}^S\) and \(\nabla _{\theta_{t}} y_{t}^T\) are mutually independent, therefore we can calculate them independently and achieve joint optimizing in an efficient way. In our experiments we set \(D_{course}=D_{exam}=D_{train}\) for efficient optimization. Though, it should be mentioned that in our framework the course data set can be unlabeled, i.e. \(D_{course} = \{x_i\}_{i=1}^N\). In this case, the cross entropy loss between the student's prediction and the ground truth is removed such that the loss for the student is exactly the knowledge distillation loss. Thus, only the supervision signal of \(D_{exam}\) is required. We leave IKD in this semi-supervised setting as future work.

\section{Experiments}
Though IKD is orthogonal to most current knowledge distillation methods to compressing large-scale PLMs such as DistilBERT~\cite{sanh2019distilbert} and TinyBERT~\cite{jiao-etal-2020-tinybert}, which distill different feature sets (such as logits, embeddings, hidden states, attention maps, etc.), it is exhaustive to test IKD with these different distillation features and their combinations. Thus, in our experiments we mainly evaluate the effectiveness of IKD based on the vanilla knowledge distillation and BERT-PKD~\cite{sun-etal-2019-patient}.

\subsection{Datasets and Models}
We conduct experiments on the General Language Understanding Evaluation (GLUE) benchmark~\cite{Wang2019GLUE}  with the backbone of BERT~\cite{devlin-etal-2019-bert}, and on ChemProt dataset~\cite{schneider-etal-2020-biobertpt} and SciCite dataset~\cite{cohan-etal-2019-structural} with backbone of SciBERT~\cite{beltagy-etal-2019-scibert}.

\paragraph{GLUE Benchmark} 
We use seven text classification tasks in the GLUE benchemark: The Corpus of Linguistic Acceptability (CoLA)~\cite{Warstadt2019CoLA}, The Stanford Sentiment Treebank (SST-2)~\cite{Socher2013Recursive}, Microsoft Research Paraphrase Corpus (MRPC)~\cite{Dolan2005MRPC}, Quora Question Pairs (QQP)~\cite{Wang2019GLUE}, Multi-Genre Natural Language Inference (MNLI)~\cite{Williams2018MNLI}, Question Natural Language Inference (QNLI)~\cite{Rajpurkar2016SQuAD} and Recognizing Textual Entailment (RTE)~\cite{Wang2019GLUE}.

\paragraph{Domain Tasks}
We use  ChemProt~\cite{schneider-etal-2020-biobertpt} and SciCite~\cite{cohan-etal-2019-structural} to evaluate the performance of our method on domain specific tasks. ChemProt consists of 1,820 PubMed abstracts with chemical-protein interactions annotated by domain experts and was used in the BioCreative VI text mining chemical-protein interactions shared task. SciCite is a dataset containing 11k annotated citation intents in biochemical and computer science domains. The standard metrics are micro F1 for ChemProt and macro F1 for SciCite. We evaluate our method on both tasks with the backbone of SciBERT~\cite{beltagy-etal-2019-scibert}.

\subsection{Experimental Setup}
Our implementation is based on PyTorch~\cite{Paszke2019pytorch}. The training and evaluation are performed on a single RTX 2080Ti or RTX 3090 GPU. For downstream tasks in GLUE we first fine-tune the teacher model for 3 epochs with learning rate of 2e-5. For domain tasks the teacher is fine-tuned for 4 epochs. The batch size is 32 for all tasks. The student model is initialized with the parameters from the first \(k\) layers of BERT base(12 layer), where \(k\) is the number of the student's layer. In our experiments we mainly evaluate our method when \(k=4\) and \(k=6\). The total number of parameters of BERT\(_{24}\), BERT\(_{12}\), BERT\(_6\) and Bert\(_4\) are 340M, 110M, 67M and 52M respectively. The batch size is set to 4 for all downstream tasks. We conduct 3 experiments for each dataset and the median is reported.

\begin{table*}[t!]
    \centering

    \begin{tabular}{l|ccccccc|c}
    \toprule
\multirow{2}{*} {Method} & MNLI & QQP  & QNLI   & SST-2 & CoLA    & MRPC     & RTE   & Macro        \\ 
            & (393k) & (364k)           & (105k)          & (67k)         & (8.5k)          & (3.7k)          & (2.5k)    & Score       \\ 

\midrule
\multicolumn{9}{c}{\textit{Dev Set}} \\
\midrule
BERT\(_{24}\) (Teacher)          & 87.0        & 88.7          & 92.6          & 93.2          & 63.4          & 88.8          & 70.0  &  83.4   \\
BERT\(_4\)-TAKD\(^\dagger\)     & 72.4        & \textbf{87}          & 82.6          & 89.1          & 34.2          & 85.2          & 59.6  & 72.9 \\
BERT\(_4\)-Annealing KD\(^\dagger\)     & 74.4        & 86.5          & 83.1          &  \textbf{89.4}    & \textbf{36.0}          & \textbf{86.2}          & 61.0  & 73.8 \\
BERT\(_4\)-FT     &   78.2    &    85.9     &      85.4    &     88.0    & 28.6          & 85.1          & 63.2  &  73.5  \\
BERT\(_4\)-KD     & 78.5        & 86.4          & 85.2          & 88.8          & 30.4          & 84.6          & 63.5  & 73.9 \\
BERT\(_4\)-IKD & \textbf{79.1}        & 86.5 & \textbf{85.5} & 89.1 & 30.9 & 86.0 & \textbf{66.4}   &  \textbf{74.7} \\
\midrule
\multicolumn{9}{c}{\textit{Test Set}} \\
\midrule
BERT\(_4\)-KD     & 78.2        & 68.8          & 84.6          & \textbf{90.1}          & 27.7          & 82.5          & \textbf{63.5}  & 70.8 \\
BERT\(_4\)-IKD & \textbf{78.4}   & \textbf{68.9} & \textbf{85.3} & \textbf{90.1} & \textbf{30.6} & \textbf{83.2} & 62.9   &  \textbf{71.3} \\
\bottomrule
\end{tabular}

    \caption{Comparisons between IKD and Vanilla KD on the dev set of GLUE tasks. The teacher network is BERT\(_{24}\) and the student network is BERT\(_{4}\). FT indicates that the student is directly fine-tuned without any soft targets. KD means that the student learns using vanilla knowledge distillation. IKD reprensents our method. The best results for each task are in-bold. Methods with \(^\dagger\) denote results reported from~\citet{Jafari2021Annealing}.} 
    \label{tab:GLUE results}
\end{table*}

\subsection{Experimental Results}
We report our results on GLUE benchmark. The results are presented in Table \ref{tab:GLUE results}. We denote BERT\(_k\) as a BERT with only \(k\) layers. BERT\(_4\)-FT is to directly fine-tune the first 4 layers of the pre-trained BERT. BERT\(_4\)-KD represents student trained with vanilla knowledge distillation loss, i.e., Eq. (\ref{eq:L_tildeA}). We also exhibit the reported result of TAKD~\cite{mirzadeh2019improved} and Annealing KD~\cite{Jafari2021Annealing} for a direct comparision. 

\paragraph{Results on GLUE Benchmark} 
As shown in Table~\ref{tab:GLUE results}, IKD surpasses the vanilla knowledge distillation method on all tasks. For RTE, IKD outperforms directly finetuning by 3.2\% and knowledge distillation by 2.9\%. For QNLI and MRPC, although the performance of knowledge distillation compared with directly fine-tuning is even detrimental, IKD can consistently improve the students. We also evaluate our method on the GLUE test server for comparision on the test set and we show the results in Table \ref{tab:GLUE results}. On the GLUE leaderboard our method consistently improves the performance.

\paragraph{Results on Domain Tasks} 
Experimental results on the ChemProt and SciCite test set are shown in Table \ref{tab:ChemProt test results}, which demonstrate the effectiveness of IKD on domain specific task. It's worth noting that on the SciCite dataset our approach is even able to make the student outperform the teacher.

\begin{table}[thbp]
\centering
\begin{tabular}{l|cc} 
\toprule
\multirow{2}{*}{Method} & ChemProt & SciCite \\ 
                & (4.2k) & (7.3k) \\
\midrule
SciBERT\(_{12}\)(Teacher) & 84.5 & 86.1\\
SciBERT\(_6\)-FT & 78.1 & 85.0\\
SciBERT\(_6\)-KD & 79.3 & 85.7\\
SciBERT\(_6\)-IKD & \textbf{79.9} & \textbf{86.6} \\
\bottomrule
\end{tabular}
\caption{Experimental results of SciBERT on the test set of ChemProt and SciCite.}
\label{tab:ChemProt test results}
\end{table}

\paragraph{Compatibility with Patient Knowledge Distillation} To evaluate the compatibility of IKD with more competitive distillation methods, we further apply IKD on BERT-PKD~\cite{sun-etal-2019-patient} (denoted as BERT-PKD-IKD) with teacher model as BERT-base (12 layers) and student as BERT\(_6\). BERT-PKD exploits information from hidden features for distillation. We follow PKD-Skip strategy which is shown to be better than PKD-Last as suggested by~\citet{sun-etal-2019-patient}. Experimental results are listed in Table~\ref{tab:BERT-PKD GLUE dev results}, which demonstrate the versatility of IKD with stronger baselines. 

\begin{table}[t]
\centering
\setlength\tabcolsep{6pt}
\begin{tabular}{l|ccccc} 
\toprule
Method  & QNLI     & SST-2       & MRPC          & RTE     \\ 
\midrule
BERT\(_{12}\)(Teacher)    & 90.4          & 92.4        & 88.9          & 68.2  \\
BERT\(_6\)-PKD  & 88.3 & 89.1 &  87.6 & 66.8   \\
BERT\(_6\)-PKD-IKD & \textbf{88.7} & \textbf{90.3} & \textbf{87.7} & \textbf{67.1}    \\
\bottomrule
\end{tabular}
    \caption{Performance based on patient knowledge distillation on GLUE dev set. PKD represents patient knowledege distillation and PKD-IKD means combination of PKD and IKD.} 
    \label{tab:BERT-PKD GLUE dev results}
\end{table}

\section{Analysis}
In this section, we take a closer look at IKD by conducting empirical analysis to provide some insights to understand how it works. In particular, we focus on the  dynamic of generating soft targets and the feedback from the student.

\subsection{About the Soft Targets}
\label{sec:anal_soft_targets}
\begin{figure*}[t!]
\begin{center}
\includegraphics[width=.85\linewidth]{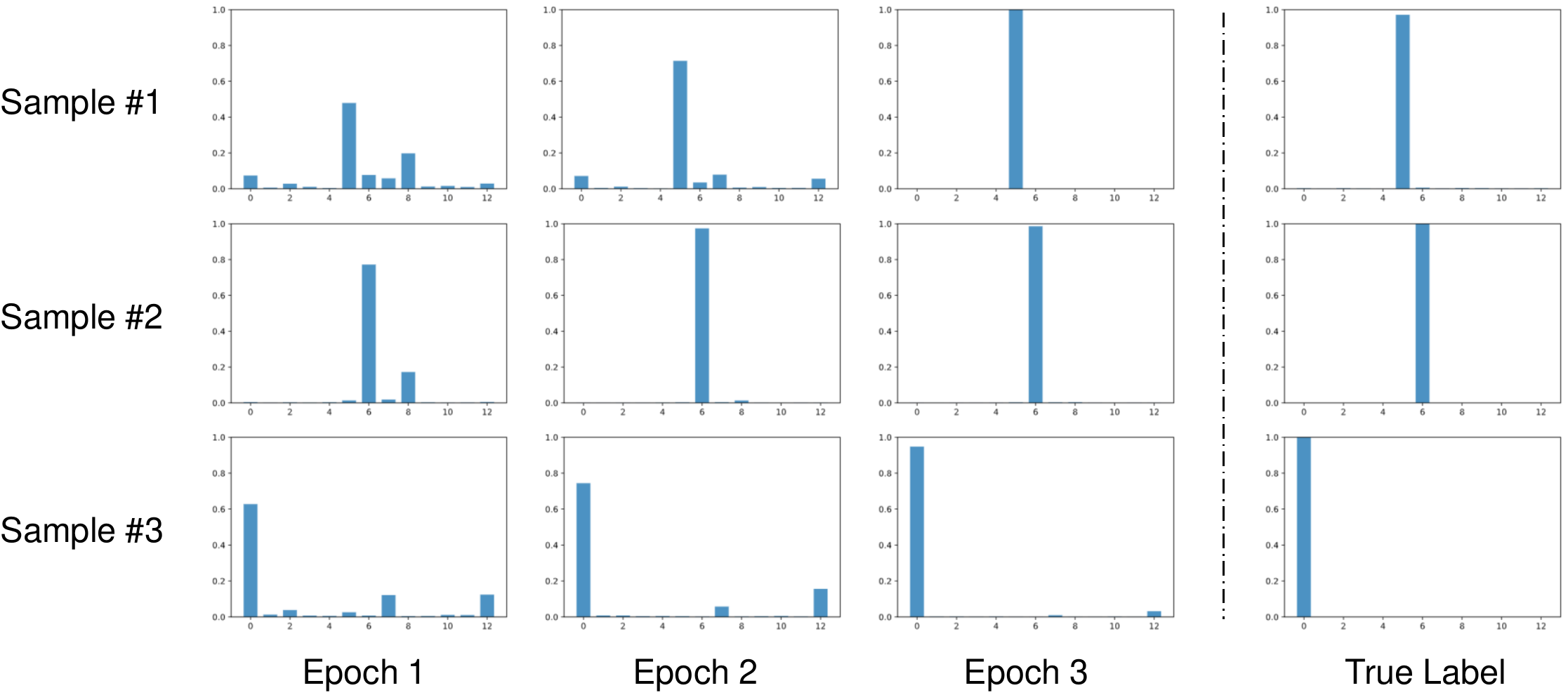}
\caption{Visualization of the generated soft targets of three training epochs on ChemProt dataset produced by the corresponding teacher. We can find that the soft targets are softer at the early stage and become sharp as the training goes on. Finally the soft targets are quite similar to the one-hot distribution.}
\label{fig:smoothness of soft targets}
\end{center}
\end{figure*}

\begin{table}[t]
\centering
\setlength\tabcolsep{6pt}
\begin{tabular}{l|ccc} 
\toprule
Dataset & Epoch 1 & Epoch 2 & Epoch 3 \\
\midrule
 MRPC & 0.2637 & 0.1076 &  0.0343 \\
 RTE & 0.4461 & 0.2313 & 0.0845  \\
 ChemProt & 0.5049 & 0.2686 & 0.1451  \\
\bottomrule
\end{tabular}
    \caption{The average entropy of output distribution for each epoch on MRPC, RTE and ChemProt dataset.} 
    \label{tab:Entropy results}
\end{table}

Since IKD allows the teacher to dynamically adjust its soft targets by using the feedback from the student, it would be interesting to see how the output of the teacher changes as the training goes on. In Figure~\ref{fig:smoothness of soft targets} we visualize the output soft targets of the teacher output in different training epochs. We can find that the soft targets are relatively smooth at the beginning and become sharper as the training progresses. This is consistent with the hand-designed smoothing schedules~\cite{Dogan2020Label,Jafari2021Annealing}, which introduce a hyper-parameter to control how quickly the soft targets converge to the one-hot distribution. In contrast, our method automatically learns to adjust the smoothness of the soft targets based on the feedback from the student. In Table \ref{tab:Entropy results} we further use entropy to measure the smoothness of output targets in different epochs on full training sets of MRPC, RTE and ChemProt. The average entropy goes down over the training process as shown, which confirms our observation as well.

\subsection{About the Feedback}
\label{sec:weight}
As mentioned in Implementation Details, the gradient of the meta loss can be viewed as a weighted sum of the gradient of teacher's prediction w.r.t. its parameters, i.e. \(\nabla_{\theta}y^T\) (See Eq. (\ref{eq:L_B approx})). The weighting vector \(\alpha \nabla _{\phi} \mathcal{L}_{meta} \cdot \nabla _{\phi} \log y^S\) can be regarded as the feedback of the student. Assume \(fb = \alpha \nabla _{\phi} \mathcal{L}_{meta} \cdot \nabla _{\phi} \log y^S\), Eq. (\ref{eq:L_B approx}) can be written as
\begin{equation}
    \nabla_{\theta}\mathcal{L}_{meta} \approx fb\cdot \nabla_{\theta}y^T =\nabla_{\theta}\sum_{i=1}^C fb_i\cdot y_{i}^T,
\end{equation}
where \(fb\in\mathbb{R}^C\), \(y^T\in\mathbb{R}^C\), \(C\) is the number of classes. 

\begin{figure}[t!]
    \centering
    \begin{subfigure}{.9\linewidth}
    \centering
    \includegraphics[scale=0.5]{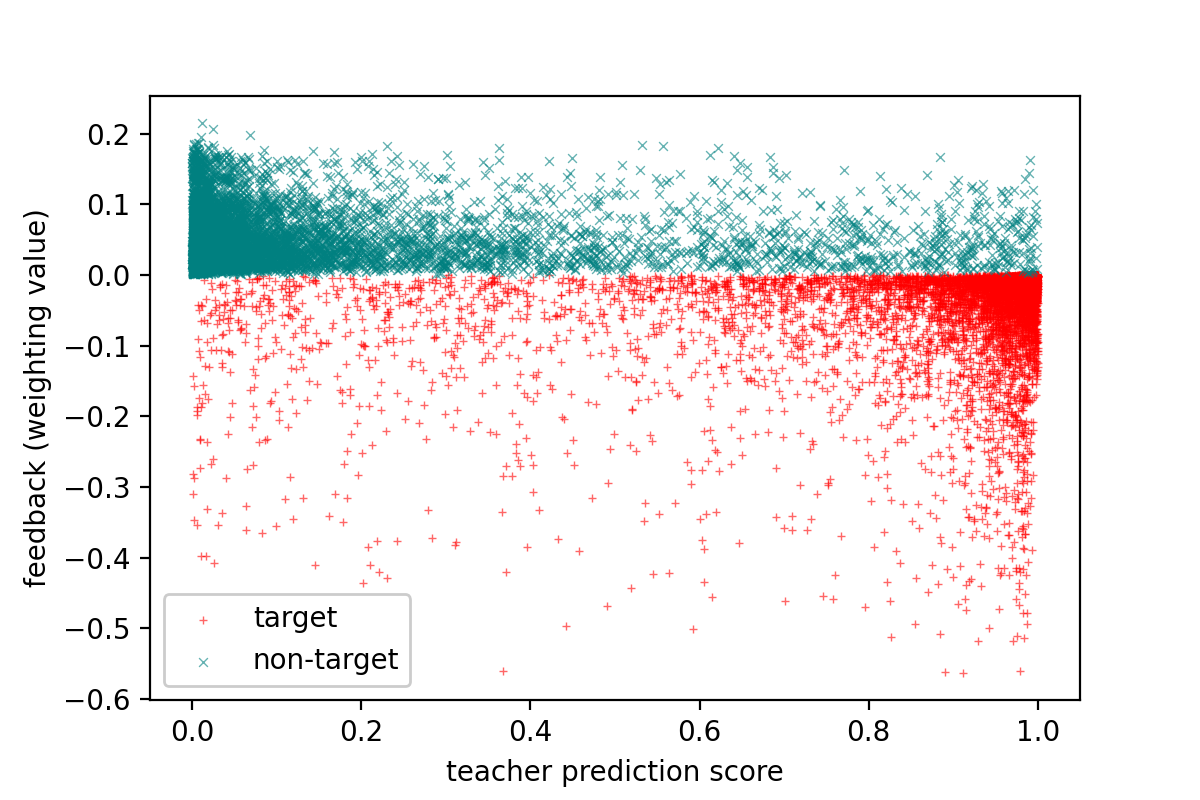}
    \caption{Feedback of target vs. non-target.}
    \label{fig:logits_weight_a}
    \end{subfigure}\hfill
    \begin{subfigure}{.9\linewidth}
    \centering
    \includegraphics[scale=0.5]{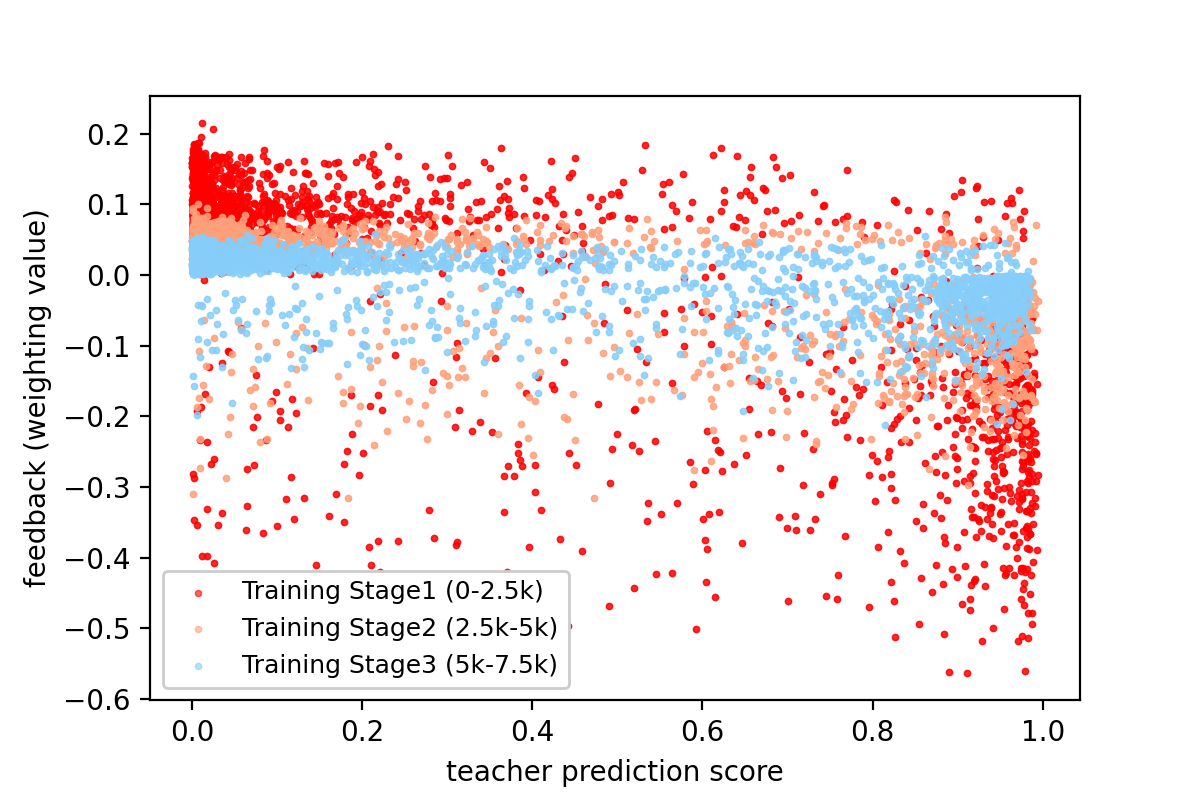}
    \caption{Feedback in different training stage.}
    \label{fig:logits_weight_b}
    \end{subfigure}\hfill
    \caption{Visualization results to show the relationship between student feedback and teacher prediction scores on CoLA dataset.}
    \label{fig:logits_weight}
\end{figure}

In this section, we attempt to analyze the relationship between the student's feedback \(fb\) and the teacher's prediction \(y^T\). In particular, we plot \((fb_i, y^T_i)\) for each \(i\in \{1,\cdots,C\}\) in Figure~\ref{fig:logits_weight}. The data points are collected from the 7,500 training steps on CoLA dataset. CoLA is a binary classification task so we can plot \(C=2\) data points for each training step such that there are 15,000 data points in total.

Note that The greater the student's feedback \(fb_i\) for class \(i\), the greater the penalty of teacher's prediction on class \(i\). As shown in Figure~\ref{fig:logits_weight_a}, the student's feedback is always non-positive for teacher's prediction on target label, which can also be revealed by the expanded form of the feedback:
\begin{equation}
\label{eq:coefficent of the target}
\alpha \nabla _{\phi} \mathcal{L}_{meta} \nabla _{\phi} \log y^S_c = -\alpha (\nabla _{\phi} \log y_{c}^S)^2,
\end{equation}
where \(c\) is the target label. Besides, we can see that the student's feedback is usually non-negative for teacher's prediction on non-target label. Thus, the teacher's prediction is always encouraged to approach the one-hot distribution where only the target label is 1, which has been verified in analysis. Also, we find that the range of feedback on target prediction is approximately three times larger than the range of feedback on non-target prediction. 

In addition, the student's feedback can be different even when the teacher's predictions are the same. To take a deeper look, we demonstrate the value of feedback of the student at different training stage in Figure~\ref{fig:logits_weight_b}. We can see that the student feedback converges to zero as the training goes, which implies that IKD works mainly at the immediate training stage.

\section{Related Work}

Our method draws on the idea of meta-learning and applies it to knowledge distillation for language understanding. Thus, there are two lines of work that are related: (a) Knowledge distillation methods for language understanding, (b) Optimization-based meta-learning. We will introduce the two lines of work in Transformer Based KD and Optimization-Based Meta-Learning, and the intersect of the both lines in Teacher-Student Framework with Meta-Learning.


\subsection{Knowledge Distillation for NLP}
\label{sec:Transformer Based KD}
Knowledge distillation (KD) was originally proposed in domains other than NLP~\cite{Bucila2006Compression,Hinton2015Distilling}. With the emergence of large-scale pre-trained language models (PLMs) such as GPT~\cite{radford2019language}, BERT~\cite{devlin-etal-2019-bert}, and RoBERTa~\cite{liu2019roberta}, KD has gain much attention in the NLP community due to its simplicity and efficiency in compressing the large PLMs. Much effort has been devoted to compress BERT via KD, including BERT-PKD~\cite{sun-etal-2019-patient}, DistilBERT~\cite{sanh2019distilbert}, MobileBERT~\cite{Sun2020Mobile}, and TinyBERT~\cite{jiao-etal-2020-tinybert}. These methods have achieved great success by exploring informative features to be distilled, such as word embeddings, attention maps, hidden states, and output logits. Different from these work, our proposed IKD focuses on the interaction between teacher and student therefore is orthogonal to these work. \citet{Gou_2021} provides a survey of online knowledge distillation where both the teacher and the student are updated simulatenously. Our IKD can be regarded as a special case of online KD.

\subsection{Optimization-Based Meta-Learning}
\label{sec:Optimization-Based Meta-Learning}

Optimization-based meta-learning intends to learn a group of initial parameters that can fast converge on a new task with only a few gradient steps, in which the learner and meta-learner share the same architecture. As a representative method, Model-Agnostic Meta-Learning (MAML) \cite{DBLP:conf/icml/FinnAL17} uses a nested loop to find an optimal set of parameters on a variety of learning tasks, such that they can adapt well after a small number of training steps. To tackle the high-order derivative, \citet{DBLP:conf/icml/FinnAL17} also propose First-Order MAML (FOMAML) to perform the first-order approximation to reduce the computation. Similar to MAML, Reptile~\cite{nichol2018fomaml} utilizes the weights in the inner loop instead of gradient to update the meta parameters and achieves competitive performance on several few-shot classification benchmarks.

\subsection{Teacher-Student Framework with Meta-Learning}
\label{sec:Teacher-Student Framework with Meta-Learning}
Meta-learning has been proved efficient when combined with the teacher-student framework. \citet{DBLP:conf/eccv/LiuRL0H20} employ meta-learning to optimize a label generator that generates dynamic soft targets for intermediate layers. Their method is proposed for self-boosting. Therefore, the teacher and the student are required to share the same architecture and parameters.
\citet{DBLP:journals/corr/abs-2012-01266} focus on training a meta-teacher possessing multi-domain knowledge by metric-based meta-learning and then teach single-domain students. Meta Pseudo Labels \cite{DBLP:journals/corr/abs-2003-10580} leverage a teacher network to generate pseudo labels on unsupervised images to enlarge the training set of the student. They aim to train a pseudo labels generator to while IKD dedicates to compress a large model into a smaller one. Thus the pseudo label generator is trained from scratch and shares same architecture with the student while in our IKD the teacher must be a larger pretrained model which leads to the impossibility of implementation of gradient substitution in MAML. Moreover, Meta Pseudo Labels generates hard labels and uses Monte Carlo approximation to obtain gradient while soft targets from the teacher in KD are necessary for passing dark knowledge to student and gradients in IKD are obtained by adopting the chain rule manually.

\section{Conclusion}
This paper proposes the Interactive Knowledge Distillation (IKD) to empower the teacher to learn to teach with student feedback. Inspired by MAML, IKD involves two optimization steps: (1) a course step for student optimization with the guidance of teacher and (2) an exam step for teacher optimization with the feedback from student. IKD shows superiority over vanilla KD on a suit of NLP tasks. By iterating the course step and exam step we can jointly optimize the teacher and student. In addition, we have discussed a semi-supervised generalization of our method, in which the course data is allowed to be unlabeled.

\bibliography{aaai22}

\begin{thebibliography}{29}
\providecommand{\natexlab}[1]{#1}

\bibitem[{Beltagy, Lo, and Cohan(2019)}]{beltagy-etal-2019-scibert}
Beltagy, I.; Lo, K.; and Cohan, A. 2019.
\newblock {S}ci{BERT}: A Pretrained Language Model for Scientific Text.
\newblock In \emph{Proceedings of the 2019 Conference on Empirical Methods in
  Natural Language Processing and the 9th International Joint Conference on
  Natural Language Processing (EMNLP-IJCNLP)}, 3615--3620. Hong Kong, China:
  Association for Computational Linguistics.

\bibitem[{Bucila, Caruana, and Niculescu{-}Mizil(2006)}]{Bucila2006Compression}
Bucila, C.; Caruana, R.; and Niculescu{-}Mizil, A. 2006.
\newblock Model compression.
\newblock In \emph{Proceedings of the Twelfth {ACM} {SIGKDD} International
  Conference on Knowledge Discovery and Data Mining, Philadelphia, PA, USA,
  August 20-23, 2006}, 535--541. {ACM}.

\bibitem[{Cohan et~al.(2019)Cohan, Ammar, van Zuylen, and
  Cady}]{cohan-etal-2019-structural}
Cohan, A.; Ammar, W.; van Zuylen, M.; and Cady, F. 2019.
\newblock Structural Scaffolds for Citation Intent Classification in Scientific
  Publications.
\newblock In \emph{Proceedings of the 2019 Conference of the North {A}merican
  Chapter of the Association for Computational Linguistics: Human Language
  Technologies, Volume 1 (Long and Short Papers)}, 3586--3596. Minneapolis,
  Minnesota: Association for Computational Linguistics.

\bibitem[{Devlin et~al.(2019)Devlin, Chang, Lee, and
  Toutanova}]{devlin-etal-2019-bert}
Devlin, J.; Chang, M.-W.; Lee, K.; and Toutanova, K. 2019.
\newblock {BERT}: Pre-training of Deep Bidirectional Transformers for Language
  Understanding.
\newblock In \emph{Proceedings of the 2019 Conference of the North {A}merican
  Chapter of the Association for Computational Linguistics: Human Language
  Technologies, Volume 1 (Long and Short Papers)}, 4171--4186. Minneapolis,
  Minnesota: Association for Computational Linguistics.

\bibitem[{Dogan et~al.(2020)Dogan, Deshmukh, Machura, and
  Igel}]{Dogan2020Label}
Dogan, {\"{U}}.; Deshmukh, A.~A.; Machura, M.; and Igel, C. 2020.
\newblock Label-Similarity Curriculum Learning.
\newblock In \emph{Computer Vision - {ECCV} 2020 - 16th European Conference,
  Glasgow, UK, August 23-28, 2020, Proceedings, Part {XXIX}}, volume 12374 of
  \emph{Lecture Notes in Computer Science}, 174--190. Springer.

\bibitem[{Dolan and Brockett(2005)}]{Dolan2005MRPC}
Dolan, W.~B.; and Brockett, C. 2005.
\newblock Automatically Constructing a Corpus of Sentential Paraphrases.
\newblock In \emph{Proceedings of the Third International Workshop on
  Paraphrasing, IWP@IJCNLP 2005, Jeju Island, Korea, October 2005, 2005}. Asian
  Federation of Natural Language Processing.

\bibitem[{Finn, Abbeel, and Levine(2017)}]{DBLP:conf/icml/FinnAL17}
Finn, C.; Abbeel, P.; and Levine, S. 2017.
\newblock Model-Agnostic Meta-Learning for Fast Adaptation of Deep Networks.
\newblock In Precup, D.; and Teh, Y.~W., eds., \emph{Proceedings of the 34th
  International Conference on Machine Learning, {ICML} 2017, Sydney, NSW,
  Australia, 6-11 August 2017}, volume~70 of \emph{Proceedings of Machine
  Learning Research}, 1126--1135. {PMLR}.

\bibitem[{Gou et~al.(2021)Gou, Yu, Maybank, and Tao}]{Gou_2021}
Gou, J.; Yu, B.; Maybank, S.~J.; and Tao, D. 2021.
\newblock Knowledge Distillation: A Survey.
\newblock \emph{International Journal of Computer Vision}, 129(6): 1789–1819.

\bibitem[{Hinton, Vinyals, and Dean(2015)}]{Hinton2015Distilling}
Hinton, G.~E.; Vinyals, O.; and Dean, J. 2015.
\newblock Distilling the Knowledge in a Neural Network.
\newblock \emph{CoRR}, abs/1503.02531.

\bibitem[{Jafari et~al.(2021)Jafari, Rezagholizadeh, Sharma, and
  Ghodsi}]{Jafari2021Annealing}
Jafari, A.; Rezagholizadeh, M.; Sharma, P.; and Ghodsi, A. 2021.
\newblock Annealing Knowledge Distillation.
\newblock In \emph{Proceedings of the 16th Conference of the European Chapter
  of the Association for Computational Linguistics: Main Volume, {EACL} 2021,
  Online, April 19 - 23, 2021}, 2493--2504. Association for Computational
  Linguistics.

\bibitem[{Jiao et~al.(2020)Jiao, Yin, Shang, Jiang, Chen, Li, Wang, and
  Liu}]{jiao-etal-2020-tinybert}
Jiao, X.; Yin, Y.; Shang, L.; Jiang, X.; Chen, X.; Li, L.; Wang, F.; and Liu,
  Q. 2020.
\newblock {T}iny{BERT}: Distilling {BERT} for Natural Language Understanding.
\newblock In \emph{Findings of the Association for Computational Linguistics:
  EMNLP 2020}, 4163--4174. Online: Association for Computational Linguistics.

\bibitem[{Liu et~al.(2020)Liu, Rao, Lu, Zhou, and
  Hsieh}]{DBLP:conf/eccv/LiuRL0H20}
Liu, B.; Rao, Y.; Lu, J.; Zhou, J.; and Hsieh, C. 2020.
\newblock MetaDistiller: Network Self-Boosting via Meta-Learned Top-Down
  Distillation.
\newblock In Vedaldi, A.; Bischof, H.; Brox, T.; and Frahm, J., eds.,
  \emph{Computer Vision - {ECCV} 2020 - 16th European Conference, Glasgow, UK,
  August 23-28, 2020, Proceedings, Part {XIV}}, volume 12359 of \emph{Lecture
  Notes in Computer Science}, 694--709. Springer.

\bibitem[{Liu et~al.(2019)Liu, Ott, Goyal, Du, Joshi, Chen, Levy, Lewis,
  Zettlemoyer, and Stoyanov}]{liu2019roberta}
Liu, Y.; Ott, M.; Goyal, N.; Du, J.; Joshi, M.; Chen, D.; Levy, O.; Lewis, M.;
  Zettlemoyer, L.; and Stoyanov, V. 2019.
\newblock RoBERTa: {A} Robustly Optimized {BERT} Pretraining Approach.
\newblock \emph{CoRR}, abs/1907.11692.

\bibitem[{Mirzadeh et~al.(2019)Mirzadeh, Farajtabar, Li, Levine, Matsukawa, and
  Ghasemzadeh}]{mirzadeh2019improved}
Mirzadeh, S.-I.; Farajtabar, M.; Li, A.; Levine, N.; Matsukawa, A.; and
  Ghasemzadeh, H. 2019.
\newblock Improved Knowledge Distillation via Teacher Assistant.
\newblock arXiv:1902.03393.

\bibitem[{Nichol, Achiam, and Schulman(2018)}]{nichol2018fomaml}
Nichol, A.; Achiam, J.; and Schulman, J. 2018.
\newblock On First-Order Meta-Learning Algorithms.
\newblock \emph{CoRR}, abs/1803.02999.

\bibitem[{Pan et~al.(2020)Pan, Wang, Qiu, Zhang, Li, and
  Huang}]{DBLP:journals/corr/abs-2012-01266}
Pan, H.; Wang, C.; Qiu, M.; Zhang, Y.; Li, Y.; and Huang, J. 2020.
\newblock Meta-KD: {A} Meta Knowledge Distillation Framework for Language Model
  Compression across Domains.
\newblock \emph{CoRR}, abs/2012.01266.

\bibitem[{Paszke et~al.(2019)Paszke, Gross, Massa, Lerer, Bradbury, Chanan,
  Killeen, Lin, Gimelshein, Antiga, Desmaison, K{\"{o}}pf, Yang, DeVito,
  Raison, Tejani, Chilamkurthy, Steiner, Fang, Bai, and
  Chintala}]{Paszke2019pytorch}
Paszke, A.; Gross, S.; Massa, F.; Lerer, A.; Bradbury, J.; Chanan, G.; Killeen,
  T.; Lin, Z.; Gimelshein, N.; Antiga, L.; Desmaison, A.; K{\"{o}}pf, A.; Yang,
  E.; DeVito, Z.; Raison, M.; Tejani, A.; Chilamkurthy, S.; Steiner, B.; Fang,
  L.; Bai, J.; and Chintala, S. 2019.
\newblock PyTorch: An Imperative Style, High-Performance Deep Learning Library.
\newblock In \emph{Advances in Neural Information Processing Systems 32: Annual
  Conference on Neural Information Processing Systems 2019, NeurIPS 2019,
  December 8-14, 2019, Vancouver, BC, Canada}, 8024--8035.

\bibitem[{Pham et~al.(2020)Pham, Xie, Dai, and
  Le}]{DBLP:journals/corr/abs-2003-10580}
Pham, H.; Xie, Q.; Dai, Z.; and Le, Q.~V. 2020.
\newblock Meta Pseudo Labels.
\newblock \emph{CoRR}, abs/2003.10580.

\bibitem[{Radford et~al.(2019)Radford, Wu, Child, Luan, Amodei, and
  Sutskever}]{radford2019language}
Radford, A.; Wu, J.; Child, R.; Luan, D.; Amodei, D.; and Sutskever, I. 2019.
\newblock Language Models are Unsupervised Multitask Learners.

\bibitem[{Rajpurkar et~al.(2016)Rajpurkar, Zhang, Lopyrev, and
  Liang}]{Rajpurkar2016SQuAD}
Rajpurkar, P.; Zhang, J.; Lopyrev, K.; and Liang, P. 2016.
\newblock SQuAD: 100, 000+ Questions for Machine Comprehension of Text.
\newblock In \emph{Proceedings of the 2016 Conference on Empirical Methods in
  Natural Language Processing, {EMNLP} 2016, Austin, Texas, USA, November 1-4,
  2016}, 2383--2392. The Association for Computational Linguistics.

\bibitem[{Sanh et~al.(2019)Sanh, Debut, Chaumond, and
  Wolf}]{sanh2019distilbert}
Sanh, V.; Debut, L.; Chaumond, J.; and Wolf, T. 2019.
\newblock {DistilBERT}, a distilled version of {BERT}: smaller, faster, cheaper
  and lighter.
\newblock \emph{arXiv preprint arXiv:1910.01108}.

\bibitem[{Schneider et~al.(2020)Schneider, de~Souza, Knafou, Oliveira, Copara,
  Gumiel, Oliveira, Paraiso, Teodoro, and
  Barra}]{schneider-etal-2020-biobertpt}
Schneider, E. T.~R.; de~Souza, J. V.~A.; Knafou, J.; Oliveira, L. E. S.~e.;
  Copara, J.; Gumiel, Y.~B.; Oliveira, L. F. A.~d.; Paraiso, E.~C.; Teodoro,
  D.; and Barra, C. M. C.~M. 2020.
\newblock {B}io{BERT}pt - A {P}ortuguese Neural Language Model for Clinical
  Named Entity Recognition.
\newblock In \emph{Proceedings of the 3rd Clinical Natural Language Processing
  Workshop}, 65--72. Online: Association for Computational Linguistics.

\bibitem[{Socher et~al.(2013)Socher, Perelygin, Wu, Chuang, Manning, Ng, and
  Potts}]{Socher2013Recursive}
Socher, R.; Perelygin, A.; Wu, J.; Chuang, J.; Manning, C.~D.; Ng, A.~Y.; and
  Potts, C. 2013.
\newblock Recursive Deep Models for Semantic Compositionality Over a Sentiment
  Treebank.
\newblock In \emph{Proceedings of the 2013 Conference on Empirical Methods in
  Natural Language Processing, {EMNLP} 2013, 18-21 October 2013, Grand Hyatt
  Seattle, Seattle, Washington, USA, {A} meeting of SIGDAT, a Special Interest
  Group of the {ACL}}, 1631--1642. {ACL}.

\bibitem[{Sun et~al.(2019)Sun, Cheng, Gan, and Liu}]{sun-etal-2019-patient}
Sun, S.; Cheng, Y.; Gan, Z.; and Liu, J. 2019.
\newblock Patient Knowledge Distillation for {BERT} Model Compression.
\newblock In \emph{Proceedings of the 2019 Conference on Empirical Methods in
  Natural Language Processing and the 9th International Joint Conference on
  Natural Language Processing (EMNLP-IJCNLP)}, 4323--4332. Hong Kong, China:
  Association for Computational Linguistics.

\bibitem[{Sun et~al.(2020)Sun, Yu, Song, Liu, Yang, and Zhou}]{Sun2020Mobile}
Sun, Z.; Yu, H.; Song, X.; Liu, R.; Yang, Y.; and Zhou, D. 2020.
\newblock MobileBERT: a Compact Task-Agnostic {BERT} for Resource-Limited
  Devices.
\newblock In \emph{Proceedings of the 58th Annual Meeting of the Association
  for Computational Linguistics, {ACL} 2020, Online, July 5-10, 2020},
  2158--2170. Association for Computational Linguistics.

\bibitem[{Vaswani et~al.(2017)Vaswani, Shazeer, Parmar, Uszkoreit, Jones,
  Gomez, Kaiser, and Polosukhin}]{vaswani2018attention}
Vaswani, A.; Shazeer, N.; Parmar, N.; Uszkoreit, J.; Jones, L.; Gomez, A.~N.;
  Kaiser, L.; and Polosukhin, I. 2017.
\newblock Attention is All you Need.
\newblock In \emph{Advances in Neural Information Processing Systems 30: Annual
  Conference on Neural Information Processing Systems 2017, December 4-9, 2017,
  Long Beach, CA, {USA}}, 5998--6008.

\bibitem[{Wang et~al.(2019)Wang, Singh, Michael, Hill, Levy, and
  Bowman}]{Wang2019GLUE}
Wang, A.; Singh, A.; Michael, J.; Hill, F.; Levy, O.; and Bowman, S.~R. 2019.
\newblock {GLUE:} {A} Multi-Task Benchmark and Analysis Platform for Natural
  Language Understanding.
\newblock In \emph{7th International Conference on Learning Representations,
  {ICLR} 2019, New Orleans, LA, USA, May 6-9, 2019}. OpenReview.net.

\bibitem[{Warstadt, Singh, and Bowman(2019)}]{Warstadt2019CoLA}
Warstadt, A.; Singh, A.; and Bowman, S.~R. 2019.
\newblock Neural Network Acceptability Judgments.
\newblock \emph{Trans. Assoc. Comput. Linguistics}, 7: 625--641.

\bibitem[{Williams, Nangia, and Bowman(2018)}]{Williams2018MNLI}
Williams, A.; Nangia, N.; and Bowman, S.~R. 2018.
\newblock A Broad-Coverage Challenge Corpus for Sentence Understanding through
  Inference.
\newblock In \emph{Proceedings of the 2018 Conference of the North American
  Chapter of the Association for Computational Linguistics: Human Language
  Technologies, {NAACL-HLT} 2018, New Orleans, Louisiana, USA, June 1-6, 2018,
  Volume 1 (Long Papers)}, 1112--1122. Association for Computational
  Linguistics.

\end{thebibliography}

\end{document}